\documentclass[runningheads]{llncs}

\RequirePackage{fix-cm}

\usepackage[noadjust]{cite}

\usepackage{graphicx}
\usepackage{epsfig}
\usepackage{dsfont}
\usepackage{amsmath}
\usepackage{amssymb}
\usepackage{boldline,multirow}

\usepackage{ulem}
\usepackage{color}
\usepackage{graphicx}
\usepackage{subfigure}
\usepackage{bbding}
\usepackage[colorlinks,
linkcolor=black,
anchorcolor=blue,
citecolor=blue
]{hyperref} 
\begin{document}
	\title{Incremental Learning for Multi-organ Segmentation with Partially Labeled Datasets} 
	%
	%
	%
	\author{Pengbo Liu\inst{1} \and
			Li Xiao\inst{1} \and
			S. Kevin Zhou\inst{1}}
	
	\authorrunning{Pengbo Liu et al.}
	%
	
	\institute{
	Institute of Computing Technology, Chinese Academy of Sciences, Beijing, China \\
		\email{\{liupengbo2019,xiaoli\}@ict.ac.cn}\\
		\email{s.kevin.zhou@gmail.com}}
	
	\maketitle              
	\begin{abstract}
		
		There exists a large number of datasets for organ segmentation, which are partially annotated, and sequentially constructed. A typical dataset is constructed at a certain time by curating medical images and annotating the organs of interest. In other words, new datasets with annotations of new organ categories are built over time. To unleash the potential behind these partially labeled, sequentially-constructed datasets, we propose to learn a multi-organ segmentation model through incremental learning (IL). In each IL stage, we lose access to the previous annotations, whose knowledge is assumingly captured by the current model, and gain the access to a new dataset with annotations of new organ categories, from which we learn to update the organ segmentation model to include the new organs.
		We give the first attempt to conjecture that the different distribution is the key reason for `catastrophic forgetting' that commonly exists in IL methods, and verify that IL has the natural adaptability to medical image scenarios. 
		Extensive experiments on five open-sourced datasets are conducted to prove the effectiveness of our method and the conjecture mentioned above. 
		
	\end{abstract}

		\section{Introduction}
	\label{intro}
	%
	%
	
\begin{figure}[t]
		\centering
		\includegraphics[width=0.7\textwidth]{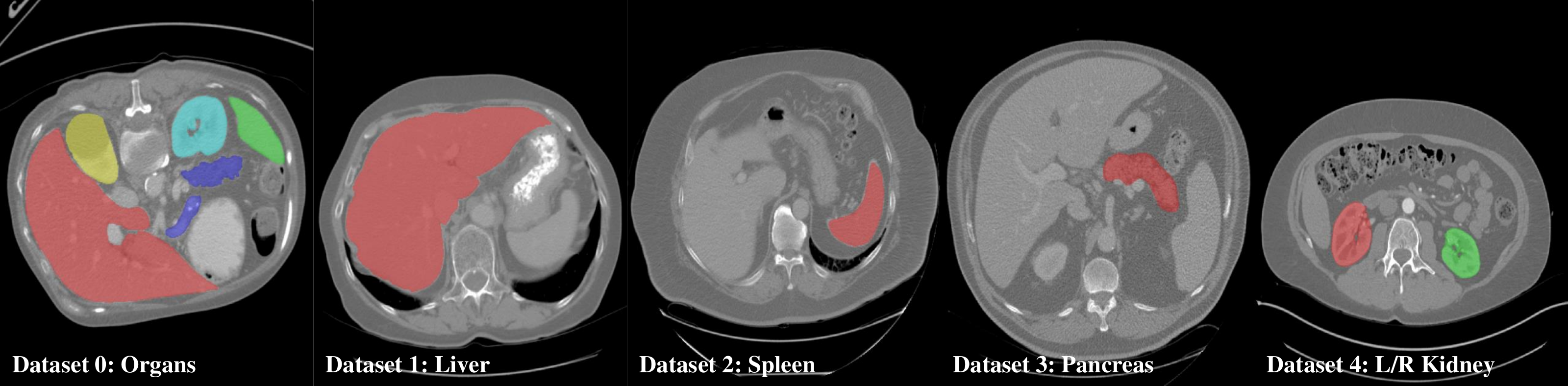}
		\caption{Five typical partially labeled CT images from five different datasets. 
		} \label{FigOrgans}
	\end{figure}

	As the performance of deep learning has been verified in the field of computer vision, a large number of supervised datasets have been open-sourced, which greatly accelerating the development of deep learning technology. Unlike natural image datasets~\cite{imagenet,coco,pascalvoc,ade20k} that are almost completely labeled for common categories, constructing a high-quality medical image dataset requires professional knowledge of different anatomical structures, so full annotation is very difficult to achieve in medical image scenarios, especially for segmentation tasks. Multi-organ segmentation is a very important task in medical image analysis scenes~\cite{zhou2019handbook,zhou2021review}. However, there exist now many partially labeled datasets~\cite{matlas,MSD,kits19_url3} that only with annotation of the organs of interest to the dataset builders. 
	Fig.~\ref{FigOrgans} gives some example images in partially labeled datasets. 
	
	There exists a `knowledge' conflict from these partially labeled datasets, \textit{e.g.}, the liver is marked as foreground in Dataset 1 and background in Datasets 2-4, as shown in Fig.~\ref{FigOrgans}. Such a conflict prevents the direct utilization of all these datasets together, which limits their potential usefulness. So far, there has been some emerging research~\cite{PIPO,PaNN,ShiMargExc} on how to mix them together, and the performance of multi-organ segmentation was improved, proving that the unlabeled data in partially labeled datasets is also helpful for learning. As clinical needs increase, more categories and labeled datasets will be added, and current methods must retrain all datasets every time. When the aggregate scale of the datasets is large, there will be great pressure on storage and efficiency.
	
	Incremental learning (IL) is a staged learning method, which learns new categories incrementally and loses access to the previous annotated images with old categories, making it an ideal choice for dealing with the above-mentioned storage and efficiency issues with better scalability in the future. And it can also solve the ethical and moral issues of sharing medical data by sharing model parameters. 
	The main challenge in IL is the so-called `catastrophic forgetting'~\cite{mccloskey1989catastrophic}: how to keep the performance on old classes while learning new ones. IL has been studied for object recognition~\cite{lwf,kirkpatrick2017overcoming,icarl2017,LUCIR2019,meta2020} and detection~\cite{9035099,8288446,shmelkov2017incremental}, but less in segmentation~\cite{sensing2019,ILT,MiB,ozdemir2019extending}. 
	In 2D medical image segmentation, Ozdemir and Goksel~\cite{ozdemir2019extending} made some attempts using the IL methods in natural images directly, with only two categories, and it mainly focuses on verifying the possibility of transferring the knowledge learned in the first category with more images to a second category with less images. In MiB~\cite{MiB}, Cermelli et al. solved knowledge conflicts existing in other IL methods~\cite{lwf,ILT} by remodeling old and new categories into background in loss functions of learning new categories and preserving old knowledge respectively, achieving a performance improvement.
	
	However, catastrophic forgetting is still obvious even though a knowledge distillation loss is used commonly for combatting it. We make a hypothesis 
    that the distillation loss can only be applied to the dataset at different stages \textit{under the same distribution}. 
	Different distributions cause the old model to output wrong responses to contents of \textit{unseen categories}  or seen categories that are quite different in appearance, violating the implicit assumption for the distillation to work. This is why `15-5' setting and `Overlapped' setting in MiB~\cite{MiB}, whose distributions in different stages are closer, perform better than other comparison settings. 

	Compared with nature images, we believe medical images are inherently adaptable to IL due to the relatively fixed anatomical structures of the human body, e.g. liver is just close to right kidney, that old categories objects will emerge in new categories learning stage. This feature can maintain the \textit{distribution consistency} of the datasets in different stages \textit{to a large extent}. Then there raises an interesting question: \textit{will the IL perform better on the medical image segmentation tasks?}
	
	In this work, we present a novel study on incremental learning models for multi-organ segmentation task with four stages to aggregate five partially labeled datasets in medical image scene. Our main contributions can be summarized as:
	\begin{itemize}
	\item We give \textit{the first attempt to perform IL on multi-organ segmentation task}, and \textit{firstly verify the effectiveness} on multiple partially annotated datasets. 
    \item Our extensive experiments on five open-sourced datasets help to prove our hypothesis that \textit{different distributions} in different stages is the key reason for catastrophic forgetting in current IL settings.
    
	\end{itemize}

	\section{Methodology}
	\begin{figure}[t]
		\centering
		\includegraphics[width=0.8\textwidth]{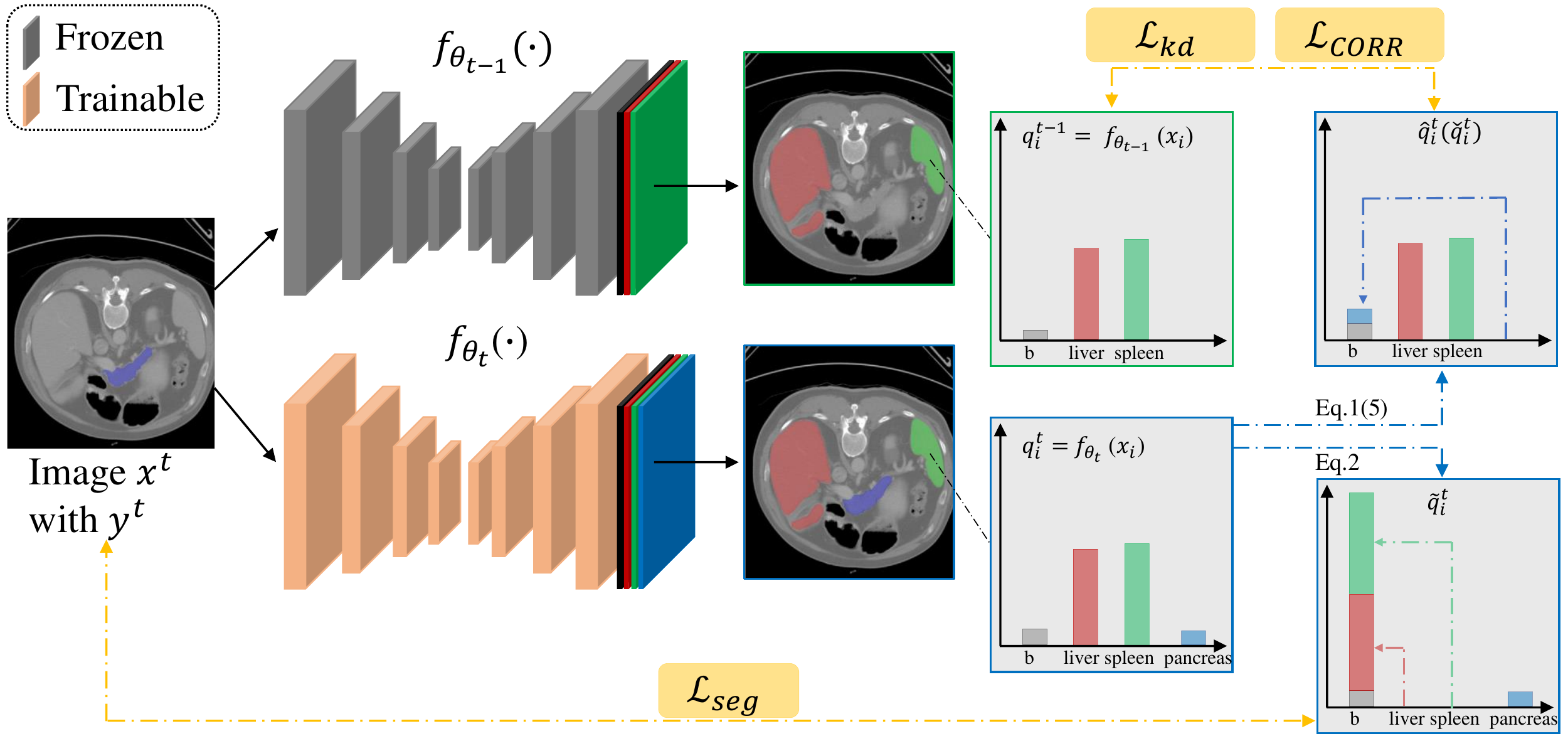}
		\caption{Overview of the $t^{th}$ stage of IL in multi-organ segmentation.} \label{Overview}
	\end{figure}
	\subsection{Problem Definition}
	
	
	The overview of $t^{th}$ stage of IL in our method is shown in Fig.~\ref{Overview}. Given an input image $x^t\in \mathcal{X}^t$, which is composed by a set of voxels $\mathcal{I}$, we firstly process it by the model in current stage, $f_{\theta_t}(\cdot)$ with trainable parameters $\theta_t$, getting the output $q^t=f_{\theta_t}(x^t)$. 
	For the learning of new categories ($\mathcal{C}^t$) in current stage, a one-hot vector $y^t$ is the ground truth for $\mathcal{C}^t$ in $x$. Label space $\mathcal{Y}^t$ is expanded from $\mathcal{Y}^{t-1}$ with $\mathcal{C}^t$, $\mathcal{Y}^t=\mathcal{Y}^{t-1} \cup \mathcal{C}^t=\mathcal{C}^0 \cup \mathcal{C}^1\cup ...\cup \mathcal{C}^t$. Note that the annotations of the old categories $\mathcal{Y}^{t-1}$ will be inaccessible in the new stage under ideal IL settings. For preserving the knowledge of old categories, we process $x$ by old model $f_{\theta_{t-1}}(\cdot)$ with frozen parameters $\theta_{t-1}$ and get $q^{t-1}=f_{\theta_{t-1}}(x^t)$ for reference. Trainable $\theta_t$ in $t^{th}$ stage is expand with $\Theta_{t}$, $\theta_t=\theta_{t-1}\cup\Theta_{t}$. We initialize $\Theta^{t}$ the same as MiB~\cite{MiB}. 
	
	\subsection{Background Remodeling}
	While the relatively fixed anatomical structure of human body in medical image brings help for IL, it also makes the `knowledge' conflict more obvious. For example, in Fig.~\ref{Overview}, the label of voxel $i$ on spleen is background in ground truth $y^t$ of $t^{th}$ stage. If we directly calculate the loss based on $q^t$ and $y^t$, it will \textit{punish the correct response} on spleen channel, and the same for other channels.  
	Different from ~\cite{ozdemir2019extending} based on Learning without Forgetting~\cite{lwf} using $q^t$ directly, we remodel the background (b) channel of $q^t$  based on MiB~\cite{MiB}
	by moving probabilities of new classes or old classes to background class, getting $\hat{q}^t$ and $\tilde{q}^t$ for the following calculation of the loss functions. Their definition is shown in 
	Eq.~\ref{eq2} and Eq.~\ref{eq3}, respectively. 
	
	\begin{equation}\label{eq2}
		\hat{q}^t_{i,c}=\left\{
		\begin{array}{lp{8mm}<{\centering}l}
			\text{exp}(q^t_{i,b}+\sum_{c\in \mathcal{C}^{t}}q^t_{i,c})/\sum_{c\in \mathcal{Y}^{t}}\text{exp}(q^t_{i,c}) & &\textit{if } c = b\\
			\text{exp}(q^t_{i,c})/\sum_{c\in \mathcal{Y}^t}\text{exp}(q^t_{i,c})&& \textit{if } c \in y^{t-1} \& c\neq b
		\end{array} \right. 
	\end{equation}
	
	\begin{equation}\label{eq3}
		\tilde{q}^t_{i,c}=\left\{
		\begin{array}{lp{8mm}<{\centering}l} 
			\text{exp}(\sum_{c\in \mathcal{Y}^{t-1}}q^t_{i,c})/\sum_{c\in \mathcal{Y}^{t}}\text{exp}(q^t_{i,c}) & &\textit{if } c = b\\
			0 && \textit{if } c \in y^{t-1} \& c\neq b\\
			\text{exp}(q^t_{i,c})/\sum_{c\in \mathcal{Y}^t}\text{exp}(q^t_{i,c}) & & \textit{if }  c \in \mathcal{C}^t
		\end{array} \right. 
	\end{equation}

	\subsection{Loss Functions}
	In the IL setting, the whole loss function $\mathcal{L}$ is composed by $\mathcal{L}_{seg}$ for learning new knowledge of new categories and $\mathcal{L}_{kd}$ for preserving old knowledge distilled from the previous model, $f_{\theta_{t-1}}$. 	
For $\mathcal{L}_{seg}$, the cross-entropy loss is the most commonly used. We also invoke Dice loss~\cite{vnet} into $\mathcal{L}_{seg}$, which is verified useful in medical image segmentation. 

	
	
	\begin{equation}
		\begin{aligned}
			\mathcal{L}&=\mathcal{L}_{seg}(\tilde{q}^t, y^t)+\mathcal{L}_{kd}(\hat{q}^t,\sigma(q^{t-1}))\\
			&=\mathcal{L}_{CE}(\tilde{q}^t, y^t)+\mathcal{L}_{Dice}(\tilde{q}^t, y^t)+\mathcal{L}_{kd}(\hat{q}^t,\sigma(q^{t-1}))
		\end{aligned}
	\end{equation}

	Where $\sigma$ is the softmax operation. 
	

	 \subsubsection{CORR Loss}
	 We also devise a new corrective (CORR) loss to \textit{reduce the low confident knowledge} and remove some false positive predictions, maybe caused by distribution disturbance between different datasets. CORR loss weakens voxel $i$ with low confident response and enhances the influence of voxel $i$ with high confident response, which is implemented by $W$ defined in Eq.~\ref{eqw}, where $C$ is the target category in voxel $i$, $\mathcal{THR}$ is the threshold of confidence and $n$ is the scale exponent. We set $\mathcal{THR}$ and $n$ to 0.95 and 12 empirically. $y^{t-1}_{pseu}=onehot(argmax_{c\in\mathcal{Y}^{t-1}}q^{t-1}_{c})$.

	 \begin{equation}\label{eqw}
	 	W_{i,c}=\left\{
	 	\begin{array}{lp{8mm}<{\centering}l} 
	 		\left(\frac{\mathcal{THR}}{\sigma(q^{t-1})_{i,c}}\cdot {y^{t-1}_{pseu}}_{i,c}\right)^{n} & &\textit{if } c = C\\
	 		1 && \textit{if } c \neq C
	 	\end{array} \right. 
	 \end{equation}
	
	 Then CORR loss can be calculated as shown in Eq.~\ref{corrlosseq}. 
	
	\begin{equation}\label{eq1}
	 	\check{q}^t_{i,c}=\left\{
	 	\begin{array}{lp{8mm}<{\centering}l}
	 		q^t_{i,b}+\sum_{c\in \mathcal{C}^{t}}q^t_{i,c}& &\textit{if } c = b\\
	 		q^t_{i,c}&& \textit{if } c \in y^{t-1} \& c\neq b
	 	\end{array} \right. 
	 \end{equation}
	 
	 \begin{equation}\label{corrlosseq}
	 	\mathcal{L}_{CORR}(q^t, q^{t-1})\!=\!\mathcal{L}_{CE}(\check{q}^t\!\cdot\! W, y^{t-1}_{pseu}) \!=\! -\frac{1}{|\mathcal{I}|}\sum_{i\in \mathcal{I}}\sum_{c\in \mathcal{Y}^{t}}{y^{t-1}_{pseu}}_{i,c}\log\left(\sigma(\check{q}^{t}_{i,c}*W_{i,c})\right)
	 \end{equation}
	
\begin{figure}[t]
	 	\centering
	 	\subfigure[Low confidence in CORR loss]{
	 		\begin{minipage}[t]{0.5\linewidth}
	 			\includegraphics[width=2.2in]{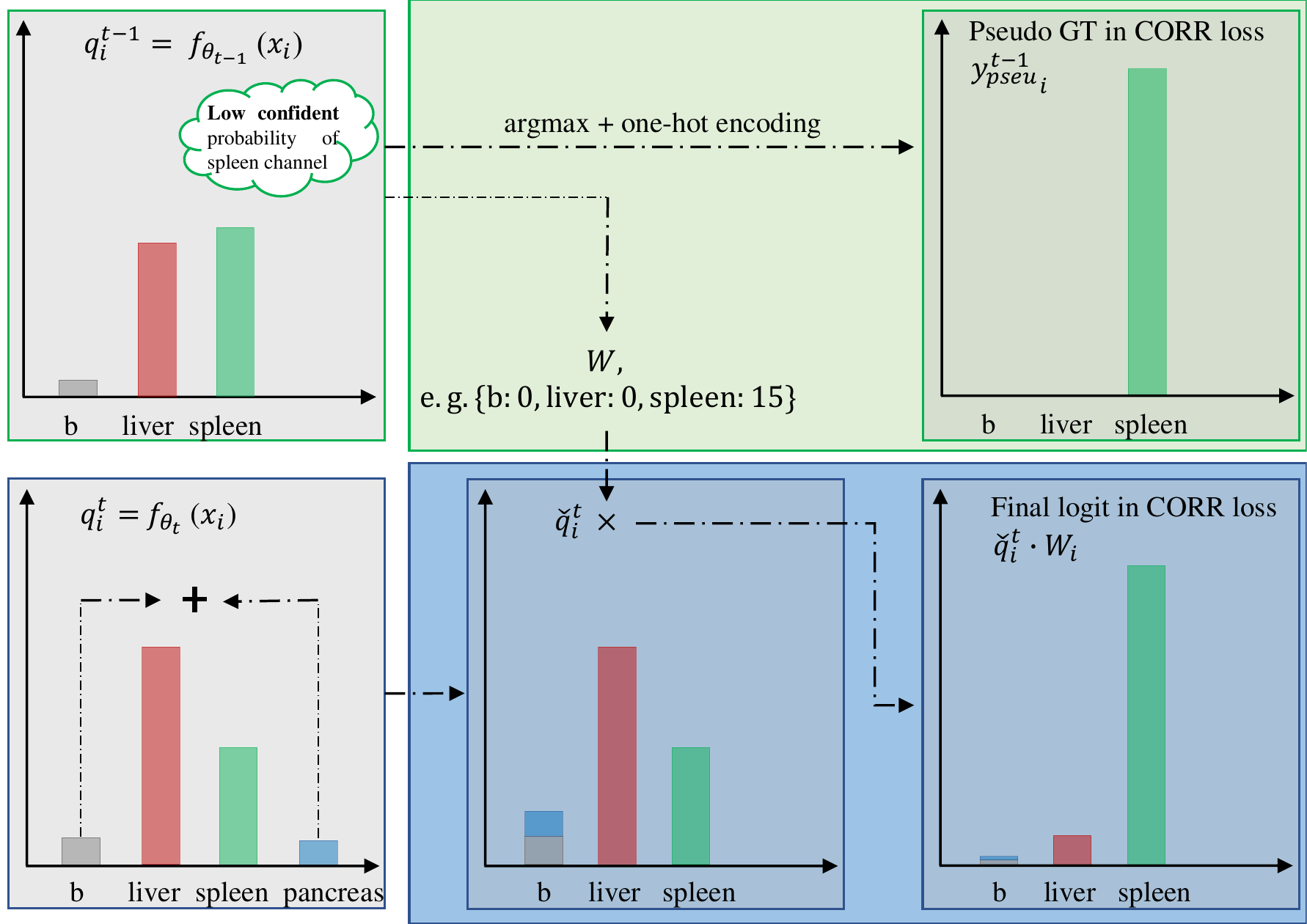}\\
	
	 		\end{minipage}%
	 	}%
	 	\subfigure[High confidence in CORR loss]{
	 		\begin{minipage}[t]{0.5\linewidth}
	 			\includegraphics[width=2.2in]{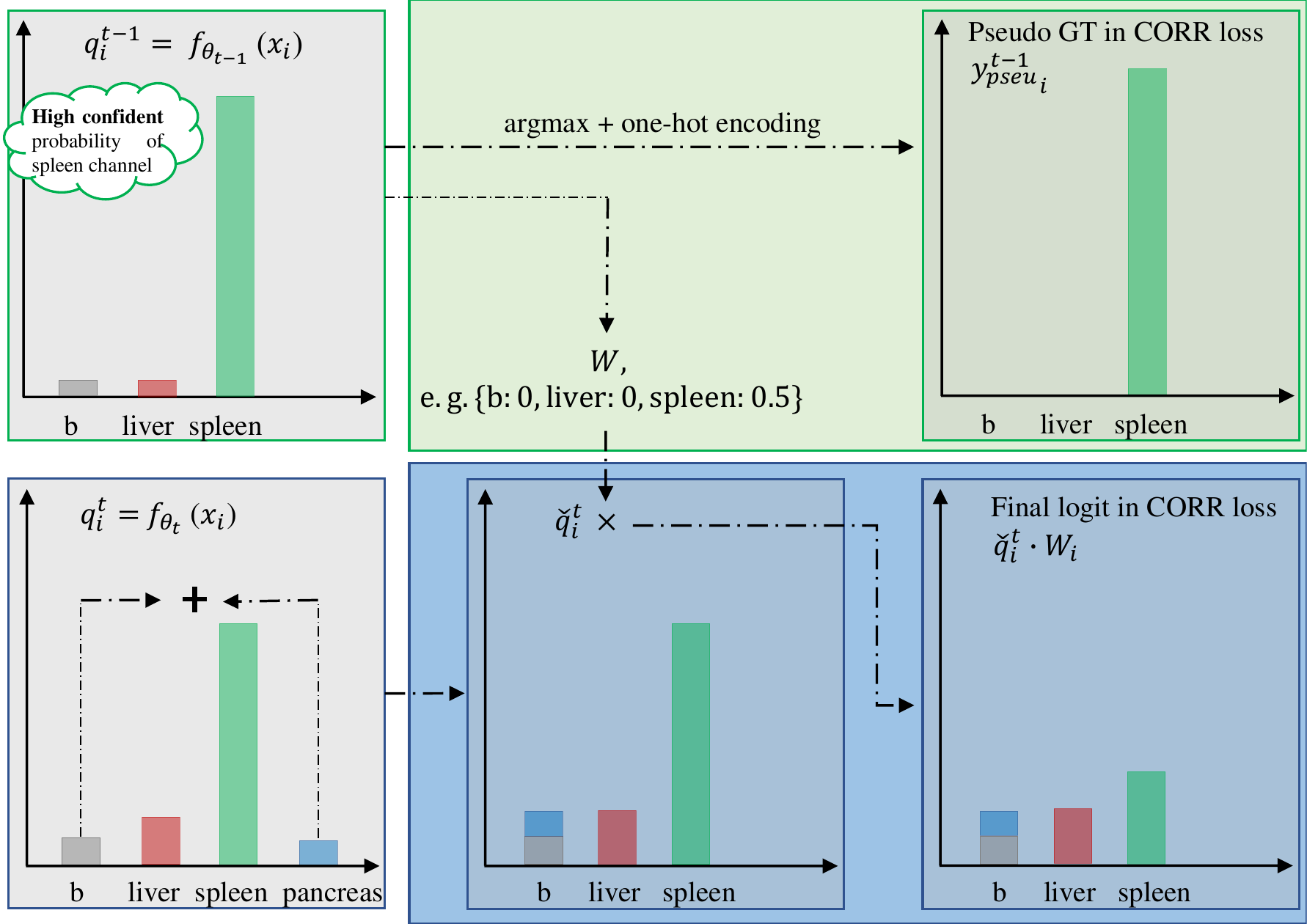}\\
	 		\end{minipage}%
	 	}
	 \centering
	 \caption{Diagram of $\mathcal{L}_{CORR}(q^t, q^{t-1})$. (a) When the confidence of the pseudo GT out of model in $stage_{i-1}$  is low, w will reduce this voxel's contribution to the CORR loss. (b) Contrary to (a) when the confidence is high.}
	 \vspace{-0.2cm}
	 \label{CORRdiagram}
	 \end{figure}

	 The diagrams of CORR loss is shown in Fig.~\ref{CORRdiagram}. In (a), When the voxel $i$ is with a low confident response from $f_{\theta_{t-1}}$, i.e. $q^{t-1}_{i,C}<\mathcal{THR}$, $q^{t}_{i,C}$ will be enlarged $W_{i,C}$ times, where $C$ is the channel of `Spleen' here. The lower confidence, the higher probability in corresponding channel and the lower contribution to loss function. Vice versa at voxels with high confidence as shown in (b). So the whole loss function in our method is shown as below, and $\omega_1$, $\omega_2$ and $\omega_3$ are set to 1, 10, 1 based on ~\cite{MiB}.

	 \begin{equation}
	 	\mathcal{L}=\omega_1\cdot\mathcal{L}_{seg}+\omega_2\cdot\mathcal{L}_{kd}+\omega_3\cdot\mathcal{L}_{CORR}
	 \end{equation}
	
	\section{Experiments}
	
	\subsection{Implementation Details}

	\subsubsection{Datasets and Preprocessing} We choose five organs in our experiments, including liver, spleen, pancreas, right kidney and left kidney, and use five CT datasets as shown in Table~\ref{summaryofdatasets}. We process all datasets to a unified spacing (1.7, 0.79, 0.79) and normalize them with mean and std of 90.9 and 65.5 respectively. We split five datasets to 5 folds and select one fold, randomly, to evaluate our method. 
	For our IL setting, five organs are collected in four stages, liver (F+$P_1$)$\rightarrow$spleen (F+$P_2$)$\rightarrow$pancreas (F+$P_3$)$\rightarrow$R/L kidney (F+$P_4$). The annotations of different organs in dataset F are used separately in our IL setting.
	%

	\subsubsection{Code} We implement our method based on open source code of 3D fullres version nnU-Net\footnote{\url{github.com/mic-dkfz/nnunet}}~\cite{FabianNNUnet_nm}. We also used MONAI\footnote{\url{https://monai.io/}} and MiB\footnote{\url{https://github.com/fcdl94/MiB}} during our algorithm development. Considering limiting the GPU memory consumption within 12Gb, our patch-size and batch-size are (80, 160, 128) and 2 in our experiments. We train the network with the same optimizer and learning rate policy as nnU-Net for about 400 epochs. The initial learning rate of the first stage and followed stages are set to 3e-4 and 15e-5.
	
	\begin{table}[t]
		\centering
		\caption{A summary of five benchmark datasets used in our experiments. [13] means 13 organs are in original dataset. We ignore other eight organs in our experiments. [T] means there are tumor labels in original dataset and we merge them into corresponding organs.}\label{summaryofdatasets}
		\newsavebox{\tablebox}
		\begin{lrbox}{\tablebox}
			\begin{tabular}{l|c|c|c|c|c}
				\hline
				Datasets &  Modality &  \# of labeled volumes & Annotated organs & Mean spacing (z, y, x) & Source\\
				\hline
				Dataset0 (F) 	 & CT & 30 	& Five organs [13] 	& (3.0, 0.76, 0.76) &Abdomen in \cite{matlas}\\
				Dataset1 ($P_1$) & CT & 131 & Liver [T]		& (1.0, 0.77, 0.77) 	&Task03 in ~\cite{MSD}\\
				Dataset2 ($P_2$) & CT & 41 	& Spleen 			& (1.6, 0.79, 0.79) &Task09 in ~\cite{MSD}\\
				Dataset3 ($P_3$) & CT & 281 & Pancreas [T] 	& (2.5, 0.80, 0.80) 		&Task07 in ~\cite{MSD}\\
				Dataset4 ($P_4$) & CT & 210 & L\&R Kidneys [T] & (0.8, 0.78, 0.78) 		&KiTS~\cite{kits19_url3}\\
				\hline	
				All				 & CT & 693 & Five organs & (1.7, 0.79, 0.79) & -\\
				\hline
			\end{tabular}
		\end{lrbox}
		\scalebox{0.82}[0.82]{\usebox{\tablebox}}
	\end{table}
	
	\begin{table}[t]
		\centering
		\caption{In the last stage, the 95$^{th}$ percentile Hausdorff distance (HD95) of the segmentation results of different methods on different datasets. The best result is shown in \textbf{bold}.}\label{HD}
		\begin{lrbox}{\tablebox}
			\begin{tabular}{l|cc|cc|cc|cc|cc|c}
				\hline
				Methods$\backslash$Organs &  Liver$\in$F &  Liver$\in$$P_1$ & Spleen$\in$F & Spleen$\in$$P_2$ & Pancreas$\in$F & Pancreas$\in$$P_3$ & R Kidney$\in$F& R Kidney$\in$$P_4$ & L Kidney$\in$F &L Kidney$\in$$P_4$ & Mean\\

				\hline
				$\phi_{F+P_1}$ (Liver) 		& $2.39\pm0.66$ & $10.81\pm25.54$  & - & - & - & - & - & - & -&-&-\\
				$\phi_{F+P_2}$ (Spleen)  		& -  & - & $1.58\pm0.41$  & $24.74\pm62.02$  & - & - & - & - & -&-&-\\
				$\phi_{F+P_3}$ (Pancreas)  	& -  & - & - & - & $23.23\pm33.60$  & $6.45\pm10.02$  & - & - & -&-&-\\
				$\phi_{F+P_4}$ (R/L Kidney)  	& -  & - & - & - & - & - & $26.49\pm54.34$  & $15.15\pm43.06$  & $30.13\pm61.08$	& $6.67\pm16.59$&$14.76$\\
				\hline
				$\phi_{F}$ (Five organs) 		& $1.58\pm0.41$  & $12.12\pm17.81$ & $1.00\pm0.00$ & $1.35\pm0.41$  & $5.39\pm3.82$  & $9.41\pm8.98$ & $1.36\pm0.40$ & $6.19\pm4.25$ &$2.27\pm1.87$ &$11.67\pm16.45$ &$5.23$\\
				\hline\hline
				
				FT				& nan  & nan &nan& nan &nan & nan & $4.85\pm2.33$ & $8.16\pm31.29$ & $3.97\pm1.66$ & \textbf{3.01$\pm$.6.71} & - \\
				LwF~\cite{lwf} 	& $2.33\pm0.48$  & $11.19\pm24.66$ & $46.11\pm96.71$ & $30.31\pm76.26$ & $.4.89\pm3.04$ & $9.33\pm13.54$ & $16.03\pm23.95$ & $35.90\pm57.56$ & $25.68\pm22.29$ & $49.63\pm54.46$ & $23.14$ \\
				ILT~\cite{ILT}  & $2.36\pm0.53$  & $11.13\pm25.34$ & $66.61\pm102.64$ & $30.59\pm76.31$ & $16.02\pm19.58$ & $10.37\pm15.08$ & $4.63\pm2.21$ & $29.34\pm56.31$ & $4.31\pm1.21$ & $21.80\pm36.70$ & $19.72$ \\

				MiB~\cite{MiB} 		& $2.56\pm0.76$  	&$11.52\pm25.03$  &\textbf{1.48$\pm$0.37}  &\textbf{29.04$\pm$72.38}  &\textbf{3.59$\pm$1.35}  & $6.76\pm9.71$ & $4.87\pm2.47$ &\textbf{.8.09$\pm$30.75}  &$3.63\pm1.35$ &$10.29\pm28.95$&$8.19$\\
				MiBOrgan(MiB+CORR) 	& \textbf{2.19$\pm$0.72} &\textbf{11.06$\pm$24.24}  &$1.96\pm0.99$  &$30.11\pm75.21$  &$3.97\pm2.14$  & \textbf{6.13$\pm$6.04} & \textbf{4.44$\pm$2.24} &$8.58\pm33.12$  &\textbf{3.04$\pm$0.91} &$5.21\pm12.16$&\textbf{7.45}\\
				\hline\hline
				MargExc MIA~\cite{ShiMargExc} 	& $2.84\pm1.53$  	&$4.04\pm2.64$  &$17.58\pm7.27$  &$1.00\pm0.09$  &$3.24\pm0.69$  & $3.96\pm3.27$ & $1.43\pm0.14$ &$1.28\pm0.07$  &$3.13\pm0.58$ &$1.68\pm0.68$&$4.02$\\
				\hline
			\end{tabular}
		\end{lrbox}
		\scalebox{0.50}[0.56]{\usebox{\tablebox}}
	\end{table}

	\begin{table}[t]
		\centering
		\caption{In the last stage, the Dice coefficient (DC) of the segmentation results of different methods on different datasets. The best result is shown in \textbf{bold}.}\label{DC}
		\begin{lrbox}{\tablebox}
			\begin{tabular}{l|cc|cc|cc|cc|cc|c}
				\hline
				Methods$\backslash$Organs &  Liver$\in$F &  Liver$\in$$P_1$ & Spleen$\in$F & Spleen$\in$$P_2$ & Pancreas$\in$F & Pancreas$\in$$P_3$ & R Kidney$\in$F& R Kidney$\in$$P_4$ & L Kidney$\in$F &L Kidney$\in$$P_4$ & Mean\\

				\hline
				$\phi_{F+P_1}$ (Liver) 		& $.958\pm.017$ 	 & $.964\pm.030$  & - & - & - & - & - & - & -&-&-\\
				$\phi_{F+P_2}$ (Spleen)  		& -  & - & $.951\pm.010$  & $.955\pm.028$  & - & - & - & - & -&-&-\\
				$\phi_{F+P_3}$ (Pancreas)  	& -  & - & - & - & $.809\pm.053$  & $.850\pm.071$  & - & - & -&-&-\\
				$\phi_{F+P_4}$ (R/L Kidney)  	& -  & - & - & - & - & - & $.917\pm.038$  & $.970\pm.027$  & $.913\pm.031$	& $.963\pm.042$&$.925$\\
				\hline
				$\phi_F$ (Five organs) 		& $.967\pm.010$  & $.948\pm.027$ & $.969\pm.007$ & $.955\pm.005$  & $.786\pm.091$  & $.704\pm.149$ & $.949\pm.016$ & $.884\pm.093$ &$.926\pm.057$ &$.825\pm.172$ &$.891$\\
				\hline\hline
				
				FT				& $.000\pm.000$  & $.000\pm.000$ & $.000\pm.000$ & $.000\pm.000$ & $.000\pm.000$ & $.000\pm.000$ & $.917\pm.016$ & \textbf{.978$\pm$.011} & $.919\pm.015$ & \textbf{.973$\pm$.021} & $.379$ \\
				LwF~\cite{lwf} 	& $.959\pm.017$  & $.961\pm.032$ & $.940\pm.021$ & $.956\pm.024$ & $.804\pm.044$ & $.807\pm.102$ & $.912\pm.016$ & $.944\pm.043$ & $.879\pm.044$ & $.900\pm.104$ & $.906$ \\
				ILT~\cite{ILT}  & $.958\pm.017$  & $.962\pm.029$ & $.937\pm.020$ & $.949\pm.035$ & $.795\pm.046$ & $.807\pm.096$ & $.913\pm.022$ & $.955\pm.039$ & $.912\pm.017$ & $.919\pm.103$ & $.911$ \\

				MiB~\cite{MiB} 		& \textbf{.961$\pm$.017}  	&$.959\pm.037$  &\textbf{.953$\pm$.015}  &\textbf{.953$\pm$.033}  &\textbf{.817$\pm$.048}  & \textbf{.819$\pm$.111} & \textbf{.918$\pm$.018} &$.972\pm.035$  &$.920\pm.016$ &$.952\pm.073$&\textbf{.922}\\
				MiBOrgan(MiB+CORR) 	& \textbf{.961$\pm$.017} &\textbf{.960$\pm$.034}  &$.950\pm.016$  &$.950\pm.035$  &$.809\pm.049$  & $.814\pm.111$ & $.917\pm.018$ &$.971\pm.028$  &\textbf{.921$\pm$.019} & $.953\pm.077$ & $.921$\\
				\hline\hline
				MargExc MIA~\cite{ShiMargExc} 	& $.969\pm.012$  	&$.957\pm.009$  &$.924\pm.009$  &$.970\pm.008$  &$.836\pm.006$  & $.808\pm.041$ & $.946\pm.012$ &$.952\pm.013$  &$.978\pm.013$ &$.972\pm.004$&$.931$\\
				\hline
			\end{tabular}
		\end{lrbox}
		\scalebox{0.54}[0.6]{\usebox{\tablebox}}
	\end{table}

	\subsubsection{Baselines}
	
	To verify the effect of IL approach in the collection of multiple partially annotated datasets, we first construct experiments on each organ separately (F+$P_i$). Dataset F has five organs meanwhile, we also do a five-class segmentation experiment on F directly (F). To handle the datasets constructed sequentially, simple fine-tuning (FT) is the most intuitive. And we compare our proposed method with some state-of-the-art (SOTA) methods, LwF~\cite{lwf}, ILT~\cite{ILT} and MiB~\cite{MiB}. For the results please refer to Sect.~\ref{resultandDis}.

		\begin{table}[t]
	\centering
	\caption{The Dice coefficient (DC) and 95$^{th}$ percentile Hausdorff distance (HD95) of the segmentation results in different stages. The best result is shown in \textbf{bold}. `-' means Not Applicable.}\label{tab1}
	\begin{lrbox}{\tablebox}
		\begin{tabular}{l|l|cccc|cccc}
			\hline
			&&&DC&&&&HD&&\\
			\cline{3-10}
			Setting & Organs$\backslash$Stages &  S0 & S1 & S2 & S3   &  S0 & S1 & S2 & S3 \\
			\hline
			FT &Liver 		&\textbf{.963$\pm$.028} & $.000\pm.000$  & $.000\pm.000$  &  $.000\pm.000$ & \textbf{9.23$\pm$23.26}  & nan & nan & nan\\
					&Spleen 	& -  & \textbf{.961$\pm$.013}  	& $.000\pm.000$  & $.000\pm.000$  &  - & \textbf{1.35$\pm$0.34} & nan &nan \\
					&Pancreas 	& -  & - 				& \textbf{.844$\pm$.091}  & $.000\pm.000$  &  - & - & \textbf{5.00$\pm$6.82} &nan\\
					&L Kidney 	& -  & - 				& - & \textbf{.970$\pm$.024}  &  - & - & - &\textbf{7.74$\pm$29.26}\\
					&R Kidney 	& -  & - 				& - & \textbf{.966$\pm$.027}  &  - & - & - & \textbf{3.13$\pm$6.30}\\
			\hline
			LwF	    &Liver 		&\textbf{.963$\pm$.028} 	& \textbf{.962$\pm$.028}  & \textbf{.962$\pm$.028}  &  \textbf{.961$\pm$.030} & \textbf{9.23$\pm$23.26}  & $9.43\pm23.20$  & \textbf{9.36$\pm$22.80} & $9.53\pm22.50$\\
			&Spleen 	& -  			& $.948\pm.024$  & $.948\pm.024$  & $.949\pm.024$  &  - 			 & $40.02\pm85.27$ & $44.40\pm86.85$&$37.08\pm85.98$ \\
			&Pancreas 	& -  			& - 			 & $.806\pm.094$  & $.807\pm.098$  &  - 			 & - 			   & $15.59\pm32.45$ &$8.90\pm12.97$\\
			&L Kidney 	& - 		 	& - 			 & - 			  & $.940\pm.042$  &  - 		 	 & - 			   & - 				&$33.36\pm54.84$\\
			&R Kidney 	& -  			& - 			 & - 			  & $.897\pm.098$  &  - 			 & - 			   & - 				& $46.57\pm52.10$\\
			\hline
			MiBOrgan&Liver 		&\textbf{.963$\pm$.028} & $.961\pm.032$  & $.961\pm.031$  &  \textbf{.961$\pm$.032} & \textbf{9.23$\pm$23.26}  & \textbf{9.29$\pm$22.36} & $14.70\pm36.24$ &\textbf{9.40$\pm$22.12}\\
			(MiB+CORR)&Spleen 	        & -  & $.949\pm.026$  	& \textbf{.950$\pm$.025}  &\textbf{.950$\pm$.029}  &  - & $43.13\pm84.72$ & \textbf{36.09$\pm$83.96}&\textbf{18.05$\pm$58.54} \\
			&Pancreas 	        & -  & - 				& $.819\pm.100$  & \textbf{.814$\pm$.107}  &  - & - & \textbf{8.80$\pm$14.28} &\textbf{5.92$\pm$5.82}\\
			&L Kidney 	        & -  & - 				& - & $.964\pm.033$  &  - & - & - &$8.05\pm30.97$\\
			&R Kidney 	        & -  & - 				& - & $.949\pm.073$  &  - & - & - & $4.94\pm11.38$\\
			\hline
		\end{tabular}
	\end{lrbox}
	\scalebox{0.65}[0.7]{\usebox{\tablebox}}
\end{table}

	\subsection{Results and Discussions}
	\label{resultandDis}

	We use Dice coefficient (DC) and 95$^{th}$ percentile Hausdorff distance (HD95) for comparing different methods.The results are shown in Table~\ref{DC} and Table~\ref{HD}. 
	
	\subsubsection{Annotations used separately:} When we do not aggregate these partially labeled data together by IL, there are some limitations in the results. Five-class segmentation model $\phi_F$ trained on `fully' annotated dataset F, has a good performance on itself, but can not generalize well to other datasets due to the scale of the dataset F. And we train four models, $\phi_{F+P_*}$, one model per organ segmentation task trained on corresponding datasets (F+$P_*$), then all datasets can be used. We can get the best performance on DC metric, but worse on HD95 metric. And this method is also poor in scalability and efficiency when the categories grow in the future.
	
	\subsubsection{Aggregating partially labeled datasets:} When we aggregate these partially labeled datasets together, the most intuitive method FT is the worst. It has no preservation about the old knowledge because there is no restraint for it. LwF and ILT perform better than model $\phi_F$, because they learn on much more data than dataset F, 554 vs 24. But wrong supervision limits the performance of LwF and ILT on these datasets, due to `knowledge' conflict. 
	
	After we remodel the background of the predictions out of $f_{\theta_t}$, MiB gets a large improvement on the performance of DC and HD all (MiB vs Lwf/ILT), obtaining a comparable performance on DC and an obvious improvement on HD95 compared to models trained separately. Solving `knowledge' conflict between different partially annotated datasets, not only the preservation of the performance on the old categories but also the learning of the new categories has been improved. Adding CORR loss, MiBOrgan gets an exchange of $0.1\%$ drop on DC for $9\%$ enhancement on HD95. It shows that CORR loss removes some low confident predictions and reduces false positive results, thereby reducing HD95 to a certain extent.
	
	\subsubsection{Compared with partially supervised method:} We also compare the result with SOTA partially supervised method, MargExc MIA~\cite{ShiMargExc}, which have access to all partially labeled datasets and annotations in one time. The results are taken from \cite{ShiMargExc}, which can be regarded as our upper-bound. The performance of our method is close, but without accessing all the training data in one time.
	
	\subsubsection{Performance on different stages:} In Table~\ref{tab1}, we also show the performance on old and new categories of models in different stages of three typical IL settings, FT, LwF, and MiBOrgan. We can observe that FT can always get the best results on the categories learned in current stage. Because FT only needs to focus on learning fully supervised new categories with a good `pretrained' base model, which is trained in former stage. MiBOrgan and LwF can preserve old knowledge and learn new knowledge meanwhile, due to the constraint from distillation loss. The more difficult task also makes the learning of new categories not as good as FT. MiBOrgan takes one step closer to the best through solving `knowledge' conflict existing in LwF. And we found no obvious forgetting problem in our medical image scene, which can help to prove our hypothesis --- \textit{distribution consistency in medical image helps retain knowledge of old categories}. This implies that IL is a suitable choice for medical image analysis.

	
	\section{Conclusion}
	To unleash the potential from a collection of partially labeled datasets in medical image scenarios, we introduce incremental learning (IL) to aggregate them by stages, which marks the first attempt in the literature to verify for a multi-organ segmentation task the extent of the key issue associated with IL  --- different distributions between IL stages may mislead the direction of learning process. The introduction of CORR loss also helps to reduce the false positive predictions by removing predictions with low confidence. IL methods have natural adaptability to medical image scenarios due to the relatively fixed anatomical structure of human body, which is an inspiration to natural image scene that introducing an external dataset containing old categories of objects under the similar distribution in the new stage will give the same effect. We believe it will be a valuable research direction in the future. Further, we plan to explore a universal segmentation model~\cite{huang2019u2net} based on IL method, containing organs from different regions, which presents a new challenge for using IL in medical image segmentation.
	
	%
	%
	%
	

	%
	%
	%
	\bibliographystyle{splncs04}
	\bibliography{Manuscript_PaperID_768}

\end{document}